\pdfoutput=1

\documentclass[11pt]{article}

\usepackage{authblk}
\usepackage[hyperref]{acl}
\usepackage{times}
\usepackage{latexsym}

\usepackage{microtype}

\usepackage{makecell}
\usepackage{textcomp}
\usepackage{multirow}
\usepackage{graphicx}
\usepackage{xcolor}
\PassOptionsToPackage{dvipsnames,svgnames,table}{xcolor}
\usepackage{array}
\usepackage{booktabs}
\definecolor{green}{RGB}{153,255,153}
\definecolor{hred}{RGB}{255,153,153}
\definecolor{darkblue}{RGB}{0,0,153}
\definecolor{teal}{RGB}{0,153,153}
\usepackage{soul}
\DeclareRobustCommand{\hlgreen}[1]{\sethlcolor{green}\hl{#1}}
\DeclareRobustCommand{\hlred}[1]{\sethlcolor{hred}\hl{#1}}
\newcommand{\hlsuff}[2]{\colorlet{hlcolor}{cyan!#1}\sethlcolor{hlcolor}\hl{#2}}
\newcommand{\hlnecc}[2]{\colorlet{hlcolor}{orange!#1}\sethlcolor{hlcolor}\hl{#2}}
\newcommand{\PreserveBackslash}[1]{\let\temp=\\#1\let\\=\temp}
\newcolumntype{C}[1]{>{\PreserveBackslash\centering}p{#1}}
\usepackage{times}
\usepackage{latexsym}
\usepackage{natbib}
\usepackage{enumitem}
\usepackage{comment}

\usepackage{tikz}

\usetikzlibrary{arrows,positioning} 
\tikzset{
    >=stealth',
    box/.style={
           rectangle,
           rounded corners,
           draw=black, very thick,
           text width=12em, 
           minimum height=2em,
           text centered},
    preds/.style={
           rectangle,
           rounded corners,
           draw=black, very thick,
           text width=5em, 
           minimum height=2em,
           text centered},
    arrow/.style={
           ->,
           thick,
           shorten <=2pt,
           shorten >=2pt,}
}

\usepackage[T1]{fontenc}

\usepackage[utf8]{inputenc}

\usepackage{amsmath}
\title{Necessity and Sufficiency for Explaining Text Classifiers: A Case Study in Hate Speech Detection}

\author{Esma Balk{\i}r, Isar Nejadgholi, Kathleen C. Fraser, and Svetlana Kiritchenko \\
  National Research Council Canada \\
  Ottawa, Canada \\
 \footnotesize \texttt{\{Esma.Balkir,Isar.Nejadgholi,Kathleen.Fraser,Svetlana.Kiritchenko\}@nrc-cnrc.gc.ca}\\
 }

\begin{document}
\maketitle

\begin{abstract}
We present a novel feature attribution method for explaining text classifiers, and analyze it in the context of hate speech detection. Although feature attribution models usually provide a single importance score for each token, we instead provide two complementary and theoretically-grounded scores -- \textit{necessity} and \textit{sufficiency} -- resulting in more informative explanations. We propose a transparent method that calculates these values by generating explicit perturbations of the input text, allowing the importance scores themselves to be explainable. We 
employ our method to explain the predictions of different hate speech detection models on the same set of curated examples from a test suite, and show that different values of necessity and sufficiency for identity terms correspond to different kinds of false positive errors, exposing sources of classifier bias against marginalized groups. 
\end{abstract}

\section{Introduction}\label{sec:introduction}



Explainability in AI (XAI) is critical in reaching various objectives during a system’s development and deployment, including debugging the system, ensuring its fairness, safety and security, and understanding and appealing its
decisions by end-users \citep{vaughan2021human,luo2021local}.

A popular class of local explanation techniques is feature attribution methods, where the aim is to provide scores for each feature according to how important that feature is for the classifier decision for a given input.
 From an intuitive perspective, one issue with feature attribution scores is that it is not always clear how to interpret the assigned importance in operational terms. Specifically, saying that a feature is `important' might translate to two different predictions. The first interpretation is that if an important feature value is changed, then the prediction will change. The second interpretation is that, as long as the feature remains, the prediction will not change. The former interpretation corresponds to the \textit{necessity} of the feature value, while the latter corresponds to its \textit{sufficiency}. 

To further illustrate the difference between necessity and sufficiency, we take an example from hate speech detection. Consider the utterance ``I hate women''. 
For a perfect model, the token `women' should have low sufficiency for a positive prediction, since merely the mention of this identity group should not trigger a hateful prediction. However, this token should have fairly high necessity, since a target identity is required for an abusive utterance to count as hate speech (e.g., ``I hate oranges'' should not be classified as hate speech). 
In this paper, we develop a method to estimate the necessity and sufficiency of each word in the input, as explanations for a binary text classifier's decisions.



Model-agnostic feature attribution methods like ours often perturb the input to be explained, obtain the predictions of the model for the perturbed instances, and aggregate the results to make conclusions about which input features are more influential on the model decision. 
When applying these methods to textual data, 
it is common to either drop the chosen tokens, or  replace them with the mask token for those models that have been trained by fine-tuning a masked language model such as BERT \citep{devlin2019bert}. However, deleting tokens raises the possibility that a large portion of the perturbed examples are not fluent, and lie well outside the data manifold. Replacing some tokens with the mask token partially remedies this issue, however it raises others. Firstly, the explanation method ceases to be truly model-agnostic. Secondly, a masked sentence is in-distribution for the pre-trained model but out-of-distribution for the fine-tuned model, because the learned manifolds deviate from those formed during pre-training in the fine-tuning step.

To avoid these problems we use a generative model to \textit{replace} tokens with most probable $n$-grams. Generating perturbations in this way ensures that the perturbed instances are close to the true data manifold. It also provides an additional layer of transparency to the user, so they can decide whether to trust the explanation by checking how reasonable 
the perturbed examples seem. 

Although supervised discriminative models rely fundamentally on correlations within the dataset, different models might rely on different correlations more or less depending on model architecture and biases, training methods, and other idiosyncrasies. 
To capture the distinction between correlations in the data and the direct causes of the prediction, we turn to the notion of \textit{interventions} from causal inference \citep{pearl2009causality}. 
Previous work employing causal definitions of necessity and sufficiency for XAI have assumed tabular data with binary or numerical features. The situation in NLP is much more complex, since each feature is a word in context, and we have no concept of `flipping' or `increasing' feature values (as in binary data and numerical data, respectively). Instead, our method generates perturbations of the input text that have high probability of being fluent while minimizing the probability that the generated text will also be a direct cause of the prediction we aim to explain.

As our application domain we choose hate speech detection, a prominent NLP task with significant social outcomes \citep{fortuna2018survey,kiritchenko2021confronting}. It has been shown that contemporary hate speech classifiers tend to learn spurious correlations, including those between identity terms and the positive (hate) class, which can result in further discrimination of already marginalized groups \citep{dixon2018measuring, park2018reducing,garg2019counterfactual}. 
We apply our explainability metrics to test classifiers’ fairness towards identity-based groups (e.g., women, Muslims). 
We show how necessity and sufficiency metrics calculated for identity terms over hateful sentences can explain the classifier’s behaviour on non-hateful statements, highlighting classifiers' tendencies to over-rely on the presence of identity terms or to ignore the characteristics of the object of abuse (e.g., protected identity groups vs. non-human entities). 

The contributions of this work are as follows:
\begin{itemize}
    \item We present the first methodology for calculating necessity and sufficiency  metrics for text data as a feature attribution method. Arguably, this dual explainability measure is more informative and allows for deeper insights into a model's inner workings than traditional single metrics.
    \item We use a generative model for producing input perturbations to avoid the out-of-distribution prediction issues that emerge with token deletion and masking techniques.
    \item To evaluate the new methodology, we apply it to the task of explaining hate speech classification, and demonstrate that it can detect and explain biases in hate speech classifiers.
\end{itemize}

We make the implementation code freely available to researchers to facilitate further advancement of explainability techniques for NLP.\footnote{\url{https://github.com/esmab/necessity-sufficiency}}

\section{Background and Related Work}

Explanations are often categorized as to whether they are for an individual prediction (local) or for the model reasoning as a whole (global), and whether the explanation generation is a part of the prediction process (self-explaining) or generated through additional post-processing (post-hoc) \citep{Guidotti2018, adadi2018peeking}. The necessity and sufficiency explanations presented here belong to the class of local explanation methods, as do most of the XAI methods applied to NLP data \citep{danilevsky2020survey}. It is also a post-hoc method to the degree that it is entirely model-agnostic: all it requires is binary predictions on provided inputs. 

There are a few classes of popular techniques for explaining natural language processing models
(see \citet{danilevsky2020survey} for a survey). 
One approach is \textit{feature attribution methods} that allocate importance scores to each feature. These can be architecture-specific  \citep{bahdanau2015neural, sundararajan2017axiomatic}, or model-agnostic  \citep{ribeiro2016should, lundberg2017unified}.


Another approach is \textit{counterfactual explanations}, which provide similar examples to the input in order to show what kinds of small differences affect the prediction of the model \cite{wu2021polyjuice,kaushik2021explaining,ribeiro2020beyond,ross2020explaining}. These contrastive examples are related to the concept of counterfactual reasoning from the causality literature, that formalizes the question: \textit{``Would the outcome have happened if this event had not occurred?''} in order to determine whether the event was a cause of the observed outcome \cite{pearl2009causality}. Counterfactual explanation methods are often targeted at certain semantic or syntactic phenomena such as negation \citep{kaushik2021explaining} or swapping objects and subjects \citep{zhang2019paws}, and hence do not guarantee that the counterfactuals cover the data distribution around the input text well. 

In this work, we combine methods from feature attribution and counterfactual generation models. This allows us to calculate scores that capture local feature importance, and provide counterfactual examples as justification for the assigned scores.  

\paragraph{Necessity and sufficiency.} These are two notions from causal analysis that capture what one intuitively expects a true cause of an event to exhibit \citep{pearl2009causality, halpern2016actual}. 
Several works have recently suggested applying necessity and sufficiency to explain model predictions. \citet{mothilal2021towards} used the \textit{actual causality} framework of \citet{halpern2016actual} to calculate necessity and sufficiency scores for tabular data. 
\citet{galhotra2021explaining} suggested an approach to capture the notions of necessity and sufficiency from probabilistic causal models \cite{pearl2009causality}. \citet{watson2021local} presented a different method for quantifying necessity and sufficiency over subsets of features. We follow the framework of probabilistic causal models, and adopt the definitions from \citet{galhotra2021explaining}.
In NLP explanations, necessity and sufficiency have been used for evaluating rationales \citep{zaidan2007using, deyoung2020eraser, mathew2021hatexplain}\footnote{The term \textit{comprehensiveness} is often used instead of \textit{necessity} in this context.}, however to the best of our knowledge, this is the first work to explore their usage for estimating feature attribution scores. 

\paragraph{The out-of-distribution problem in feature attribution models.} Virtually all model-agnostic feature attribution models calculate importance scores by perturbing input features and assign importance according to which feature changes the outcome the most. 
However, an issue has been raised that these perturbed inputs are no longer drawn from the data distribution that the model would naturally encounter for a given task \citep{fong2017interpretable, chang2018explaining, hooker2019a, janzing2020feature, hase2021the}. This is problematic because then, any change in the model predictions could be caused by the distribution shift rather than the removal of feature values \citep{hooker2019a}. Recently, \citet{hase2021the} have argued that the problem is due to \textit{social misalignment} \citep{jacovi2021aligning}, where the information communicated by the model differs in non-intuitive ways from the information people expect. 


One solution to address these issues is to calculate importance scores by marginalizing over counterfactuals that respect the data distribution.  \citet{kim2020interpretation} and \citet{harbecke2020considering} adopted this approach and targeted text data specifically by marginalizing over infills generated by BERT. 
In our preliminary experiments, this resulted in the model putting an overwhelmingly high probability mass to one or few very common words, making the generated perturbations relatively non-diverse.\footnote{Making the softmax scores more distributed across the vocabulary results in unpredictably disfluent infills.} As \citet{pham2021double} also pointed out, BERT is very good at guessing the masked word, doing so correctly about half of the time. This behaviour results in assigning low importance to highly predictable words regardless of their true importance. 

For this reason, we choose to use a generative language model to infill masked sections with $n$-grams. Our mask-and-infill approach is similar to that of \citet{wu2021polyjuice} and \citet{ross2021explaining}, who used fine-tuned causal language models to infill masked sections of text with variable length sequences. \citeauthor{ross2021explaining} also used the contrasting label to condition the generative model. However, both these works aim to find counterfactual examples as explanations, while we marginalize over them to calculate necessity and sufficiency of each token.

\section{Our Method}

A central idea in causal inference is that of \textit{intervention}, where a random variable is intervened on and set to a certain value. The intuition is that, if a random variable is the cause of another, then intervening on the first one should affect the other, whereas if they are correlated by other means then the intervention should not have an effect. 

\paragraph{Necessity.} Let $X \leftarrow a$ denote that the random variable $X$ has been intervened so that $X = a$. When talking about a feature vector $x$, we will denote by $x_{i \leftarrow a}$ that we intervene on the $i$th feature value and set it to $a$. For an input with features $x$ where $x_i = a$,  the necessity of $x_i = a$ for the model prediction $f(x) = y$ is defined as: 
\begin{align*}
     \mathit{N_{x_i, y}} {\small=} P_{c\sim \mathcal{D}_n(x)} (f(c_{i \leftarrow a'}) {\small=} y' | c_i = a, f(c) = y)
\end{align*}
where $a'$ is an alternative feature value such that $a' \neq a$ and $y'$ is an alternate outcome such that $y' \neq y$. $\mathcal{D}_{n}(x)$ is a distribution that covers the neighborhood of $x$, and can be defined according to the data and the implementation. In words, $x_i = a$ has high necessity for the prediction $y$ if, for those points in the neighborhood of $x$ that also have the value $a$ for the $i$th feature and the same model prediction $y$, changing the $i$th feature value from $a$ to $a'$ changes the prediction from $y$ to $y'$ with high probability.

\paragraph{Sufficiency.} The sufficiency of $x_i = a$ for the model prediction $f(x) = y$ is defined as: 
\begin{align*}
    \mathit{S_{x_i, y}} {\small=} P_{c\sim \mathcal{D}_{s}(x)}(f(c_{i \leftarrow a}) = y | c_i = a', f(c) = y')
\end{align*}
This means that if $x_i = a$ has high sufficiency for the outcome $y$, then for inputs in the neighborhood of $x$ that differ in the $i$th feature value, changing $i$th feature value to that of $a$ will flip the prediction to $f(x) = y$.

\paragraph{Interventions.} Previous works applying notions of necessity and sufficiency from causal inference to XAI assume tabular data. This makes it relatively straightforward to apply these measures to the features since \textit{a)} it is clear how to assess and compare the $i$th feature of each input and \textit{b)} there is little ambiguity in how to change one feature value to another. Both these are issues for NLP data, where each feature is a token in the context of the wider text. 

We argue that the replacements should reflect the likelihood of natural data, but should still be distinct from purely observational correlations in task-specific aspects. To achieve this balance, we sample the replacement values $a'$ conditioned both on the other parts of the text and on the opposite class $y'$. If there are two features $x_i = a$ and $x_j = b$ that are both correlated with the outcome $y$, the intervention $x_{i\leftarrow a'}$, where $a'$ is sampled in this way results in $a'$ being still plausible with respect to the context $x_j = b$, but removes the potential indirect effect that $x_j = b$ causes $x_i = a$, which causes $f(x) = y$. This allows us to distinguish which of the correlated features the model relies on more for a given prediction.  

\paragraph{Estimation.} The formulae for necessity and sufficiency suggest a naive implementation of sampling first from the neighborhood of the input, picking those samples that conform to the conditions, and intervening on the feature of interest and marginalizing over the model predictions to calculate the final value. To perform these steps for each token in a
sentence is prohibitively expensive. We therefore perform interventions on subsets of tokens at once, so that one perturbation can be used in the necessity and sufficiency estimation of multiple tokens. 


We estimate the necessity of a token by perturbing subsets of tokens containing the given token and calculating the average change in model prediction, weighted according to the size of the subset. For calculating 
necessity, we marginalize over $f(c_{i \leftarrow a'})$ where $c = x_{j_1 \leftarrow b_1, \cdots j_k \leftarrow b_k}$ for a 
random subset of features $x_{j_1}, \cdots, x_{j_k}$, not including the original feature $i$. This means that in our implementation, $\mathcal{D}_n(x)$ is an interventional distribution around $x$ rather than an observational one. In practice, we estimate a simplified version of this value where we do not explicitly condition on $f(c) = y$ in order to perform the estimation efficiently.


We consider the instances where only one or a few tokens are perturbed to have higher probability in $\mathcal{D}_n(x)$. 
As such, the weight assigned to a sample with $k$ perturbed tokens is proportional to $1/k$. This means that the difference between the original and the perturbed instance is attributed to each perturbed token equally. 

For estimating sufficiency we take the dual approach. We perturb subsets of tokens \textit{excluding} the target token, and calculate the difference between the weighted average of the model predictions and the baseline prediction. Here too, $\mathcal{D}_s(x)$ is an interventional distribution where each sample $c = x_{i \leftarrow a', j_1 \leftarrow b_1, \cdots j_k \leftarrow b_k}$ for the focus feature $x_i$ and a subset of other features $x_{j_1}, \cdots, x_{j_k}$. Even though we do not explicitly condition on $f(c) = y'$, $\mathcal{D}_s(x)$ is biased towards such $c$ because the interventions are conditioned on $y'$.
For a sequence of length $n$, the weight assigned to an instance where $k$ tokens are perturbed is $1/(n-k)$. This means that for an perturbed example that contains only a single token from the original instance, the difference from the baseline will be attributed entirely to that token, whereas if there is $k$ original tokens, the attribution is shared between them. Note that $\mathcal{D}_s(x)$ still assigns a higher probability mass to instances closer to $x$, but is less peaked than $\mathcal{D}_n(x)$.

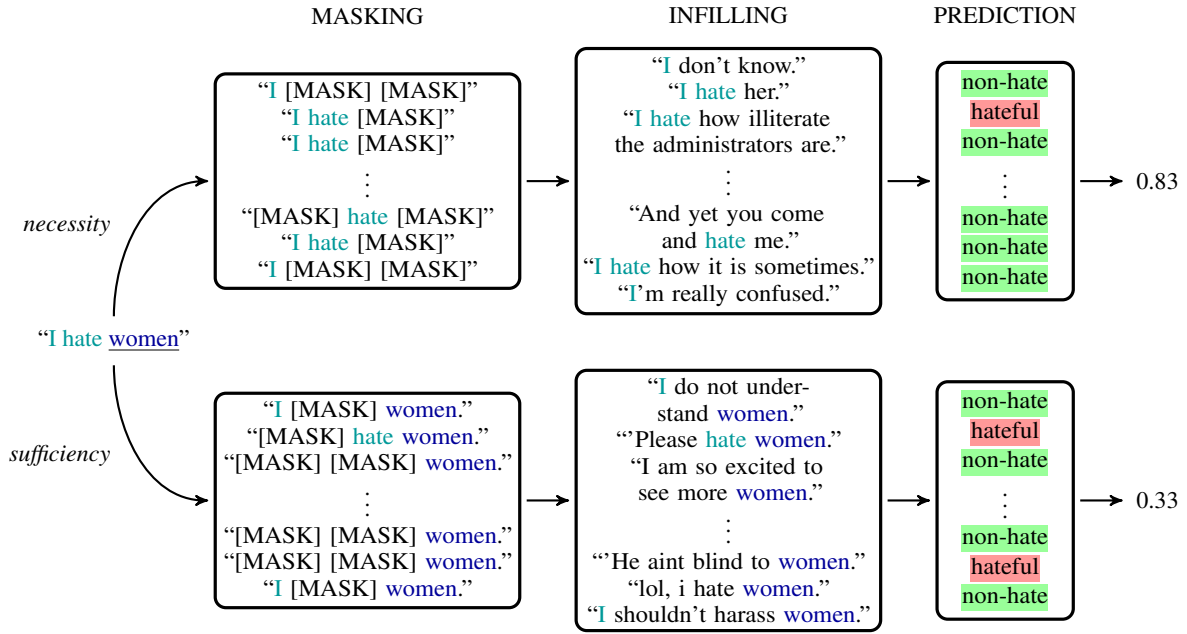
\begin{figure*}
    \centering
    \small
    \begin{tikzpicture}
    \node (sentence) {``\textcolor{teal}{I hate} \underline{\textcolor{darkblue}{women}}''};
    \node[above=1.75cm of sentence] (dummyabove) {};
    \node[below=1.75cm of sentence] (dummybelow) {};
    \node[box, right=1.2cm of dummyabove] (box11) {``\textcolor{teal}{I} {[}MASK{]} {[}MASK{]}'' \\ ``\textcolor{teal}{I hate} {[}MASK{]}'' \\ ``\textcolor{teal}{I hate} {[}MASK{]}'' \\ $\vdots$\\``{[}MASK{]} \textcolor{teal}{hate} {[}MASK{]}''  \\ ``\textcolor{teal}{I hate} {[}MASK{]}'' \\ ``\textcolor{teal}{I} {[}MASK{]} {[}MASK{]}''};
    \node[box, right=0.7cm of box11] (box12) {``\textcolor{teal}{I} don't know.'' \\ ``\textcolor{teal}{I hate} her.'' \\ ``\textcolor{teal}{I hate} how illiterate the administrators are.'' \\ $\vdots$\\ ``And yet you come and \textcolor{teal}{hate} me.''  \\ ``\textcolor{teal}{I hate} how it is sometimes.'' \\ ``\textcolor{teal}{I}'m really confused.''};
    \node[box, right=1.2cm of dummybelow] (box21) {``\textcolor{teal}{I} {[}MASK{]} \textcolor{darkblue}{women}.'' \\ ``{[}MASK{]} \textcolor{teal}{hate} \textcolor{darkblue}{women}.'' \\ ``{[}MASK{]} {[}MASK{]} \textcolor{darkblue}{women}.''\\ $\vdots$\\ ``{[}MASK{]} {[}MASK{]} \textcolor{darkblue}{women}.''  \\ ``{[}MASK{]} {[}MASK{]} \textcolor{darkblue}{women}.'' \\ ``\textcolor{teal}{I} {[}MASK{]} \textcolor{darkblue}{women}.''};
    \node[box, right=0.7cm of box21] (box22) {``\textcolor{teal}{I} do not understand \textcolor{darkblue}{women}.'' \\ ``'Please \textcolor{teal}{hate} \textcolor{darkblue}{women}.'' \\ ``I am so excited to see more \textcolor{darkblue}{women}.''\\ $\vdots$\\ ``'He aint blind to \textcolor{darkblue}{women}.''  \\ ``lol, i hate \textcolor{darkblue}{women}.'' \\ ``\textcolor{teal}{I} shouldn't harass \textcolor{darkblue}{women}.''};
    \node[preds, right=0.7cm of box12] (box13) {\hlgreen{non-hate} \\ \hlred{hateful} \\ \hlgreen{non-hate} \\ $\vdots$ \\ \hlgreen{non-hate} \\  \hlgreen{non-hate} \\ \hlgreen{non-hate} };
    \node[preds, right=0.7cm of box22] (box23) {\hlgreen{non-hate} \\  \hlred{hateful} \\ \hlgreen{non-hate} \\ $\vdots$ \\ \hlgreen{non-hate} \\ \hlred{hateful}  \\ \hlgreen{non-hate}  };
   
    \node[above=0.53cm of box11](masking){MASKING};
    \node[above=0.2cm of box12](infilling){INFILLING};
    \node[above=0.39cm of box13](prediction){PREDICTION};
    \node[right=0.7cm of box13] (necc) {0.83};
    \node[right=0.7cm of box23] (suff) {0.33};
    \draw[arrow,in=180,out=90] (sentence.north) to node[above, left=0.25cm] {\textit{necessity}} (box11.west); 
    \draw[arrow,in=180,out=-90] (sentence.south) to node[below, left=0.25cm] {\textit{sufficiency}} (box21.west); 
    \draw[arrow] (box11.east) to (box12.west); 
    \draw[arrow] (box21.east) to (box22.west);
    \draw[arrow] (box12.east) to (box13.west);
    \draw[arrow] (box22.east) to (box23.west);
    \draw[arrow] (box13.east) to (necc.west);
    \draw[arrow] (box23.east) to (suff.west);
\end{tikzpicture}

    \caption{An illustration of how necessity and sufficiency are calculated for a chosen token ``\underline{women}'' in the input ``I hate women'' that the model classifies as hateful. In the MASKING step, the subsets of tokens are masked. For the necessity calculation the masked tokens always include the focus word, and for sufficiency they always exclude it.  In the INFILLING step, the generative language model is used to infill the masked sections with $n$-grams of various lengths. These are then passed to the classifier. The necessity is the proportion of instances where changing the token changes the prediction, and sufficiency is the proportion of instances where changing other tokens does not change the original prediction. The infills are real examples generated by our method.}
    \label{fig:perturbations}
\end{figure*}




\section{Explaining Hate Speech Models}

For tasks with very skewed class distributions such as those for binary hate speech classification, it is not intuitive to ask for explanations for the majority class predictions: it is difficult to answer why a regular utterance such as ``I would like some coffee." is \textit{not} hate speech. This echoes the argument of \citet{miller2019explanation} that humans demand explanations only for selective and surprising aspects of an occurrence. We assume that there exists a majority, `neutral' class, and aim to provide explanations only for the minority, `positive' class. 

At the core of our approach is sampling replacements $x_i \leftarrow a'$ that are interventional with respect to the task, but also have high likelihood in the context of other tokens. 
We implement such a perturbation model for explaining positive predictions of hate speech classifiers, and explore what information necessity and sufficiency provide for this task through quantitative and qualitative analysis. Our implementation is \textit{task specific}, since we sample from the data labelled as `neutral' only; however, it is not \textit{dataset specific} in that it allows comparing hate speech classifiers that are trained on different datasets. 
Figure \ref{fig:perturbations} illustrates the calculations of necessity and sufficiency for an example classifier.


\begin{table*}
\centering
\small
\resizebox{\textwidth}{!}{%
\renewcommand{\arraystretch}{1.27}
\begin{tabular}{p{0.023\textwidth}m{0.28\textwidth}m{0.3\textwidth}C{0.1\textwidth}C{0.03\textwidth}}
\toprule
 & \textbf{Functionality} & \textbf{Example Test Case} & \textbf{Gold Label} & \textbf{n}  \\
\midrule
\multirow{4}{*}{\rotatebox[origin=c]{90}{\parbox[c]{2.0cm}{\centering Explicit}}} & \textbf{F1}: Expression of strong negative emotions & ``I hate {[}IDENTITY{]}.'' \newline ``You {[}IDENTITY{]} make me sick.'' & \hlred{ hateful } & 40  \\
 & \textbf{F2}: Description using very negative attributes & ``{[}IDENTITY{]} are disgusting.'' \newline ``All {[}IDENTITY{]} like them are vile.'' & \hlred{ hateful } & 40 \\
& \textbf{F3}: Dehumanisation  & ``{[}IDENTITY{]} are scum.'' \newline ``{[}IDENTITY{]} are a pest.'' & \hlred{ hateful } & 40  \\
\midrule

\multirow{2}{*}{\rotatebox[origin=c]{90}{\parbox[c]{1.25cm}{\centering Non-hate grp. ident.}}} & \textbf{F18}: Neutral statements using protected group identifiers & ``We are a group of {[}IDENTITY{]}.'' \newline ``I live with two {[}IDENTITY{]}.'' & \hlgreen{ non-hate } & 36  \\
& \textbf{F19}: Positive statements using protected group identifiers & ``I love {[}IDENTITY{]}.'' \newline ``{[}IDENTITY{]} are great.'' & \hlgreen{ non-hate } & 54 \\
\midrule

\multirow{3}{*}{\rotatebox[origin=c]{90}{\parbox[c]{2.1cm}{\centering Abuse against non-prot. targets}}} & \textbf{F22}: Abuse targeted at objects & ``I really can't stand cauliflower.'' \newline ``All cocktails like these are vile'' & \hlgreen{ non-hate } & 65 \\
& \textbf{F23}: Abuse targeted at individuals (not as member of a prot. group) & ``You make me sick.'' \newline ``I will hurt you.'' & \hlgreen{ non-hate } & 65 \\
& \textbf{F24}: Abuse targeted at non-protected groups (e.g. professions) & ``Artists are parasites to our society.'' \newline ``Death to all capitalists.'' & \hlgreen{ non-hate } & 62  \\

\bottomrule
\end{tabular}}
\caption{\label{tab:hatecheck-cases}
Test cases of interest from HateCheck \citep{rottger2021hatecheck}. We hypothesize that for identity terms in explicit test cases, if sufficiency is high then the model will have a high error rate on the test cases that capture \textit{non-hate group identity mentions}, and if necessity is low then the model will have a high error rate on the test cases that cover \textit{abuse against non-protected targets}.}
\end{table*}

\subsection{Generating perturbations} 

We use the infilling language model (ILM) of \citet{donahue2020enabling} to generate the perturbed examples. This model fine-tunes GPT-2 \citep{radford2019language} to allow infilling masked sections of text with $n$-grams, where the length of the infill varies from 1 to 7 tokens.

We fine-tune the ILM on training data that is labelled `neutral', which is the opposite of the model prediction we aim to explain. Training the perturbation model only on the neutral examples allows us to distinguish direct causes of the model prediction from correlations in data. To see this, consider the case where we would like to determine whether a given identity mention is sufficient for a positive prediction. If the ILM captures the entire data distribution rather than just the neutral class, then a large chunk of the perturbed instances might still be hateful utterances. In those cases a good model should predict the positive class even though the occurrence of the identity term is arguably not the direct cause of the prediction. Rather, the presence of the identity term causes the ILM to generate a hateful infill, which then causes the positive prediction. Using the non-hateful data distribution to train the infilling model helps avoid such cases, and enables the method to attribute importance to a token only when the classifier relies on it directly. 

To train our infilling language model, we choose four widely-used datasets for hate speech and abusive language detection that are from various sources such as Twitter \citep{founta2018large}, Reddit \citep{vidgen2021introducing}, Wikipedia comments \citep{wulczyn2017ex} and news article comments \citep{borkan2019nuanced}. For all these datasets, we 
fine-tune our generative model on benign instances from the training sets. Details for training and datasets can be found in Appendix \ref{app:ILM}.



\begin{figure*}
    \centering
    \begin{tabular}{c c c c c c | c c c c c c c}
        \hlnecc{81}{These} & \hlnecc{99}{women} &  \hlnecc{96}{disgust} & \hlnecc{45}{me} & \hlnecc{50}{so} & \hlnecc{49}{much.}  &  
        \hlnecc{33}{These} & \hlnecc{100}{Muslims} & \hlnecc{20}{disgust} & \hlnecc{0}{me} & \hlnecc{2}{so} & \hlnecc{1}{much.} \\
        \small{.81} & \small{.99} & \small{.96} & \small{.45} & \small{.50} & \small{.49} &
        \small{.33} & \small{1} & \small{.2} & \small{0} & \small{.02} & \small{.01}
        \\[5pt]
        \hlsuff{21}{These} & \hlsuff{30}{women} &  \hlsuff{37}{disgust} & \hlsuff{0}{me} & \hlsuff{11}{so} & \hlsuff{3}{much.}  &  \hlsuff{31}{These} & \hlsuff{100}{Muslims} & \hlsuff{33}{disgust} & \hlsuff{29}{me} & \hlsuff{30}{so} & \hlsuff{37}{much.}  \\
        \small{.21} & \small{.3} & \small{.37} & \small{0} & \small{.11} & \small{.03} & \small{.31} & \small{1} & \small{.33} & \small{.29} & \small{.3} & \small{.37} \\[5pt]
    \end{tabular}
    \caption{Visualization for \hlnecc{50}{necessity} and \hlsuff{50}{sufficiency} scores on an example HateCheck test case targeting \textit{women} and \textit{Muslims} for the classifier \textit{Founta2018-hate}. Darker shades correspond to higher values.}
    \label{fig:example-necc-suff}
\end{figure*}

\section{Experiments}\label{sec:experiments}

For our experiments, we focus on investigating the necessity and sufficiency of identity mentions for a sentence to be detected as hate speech, since a known bias in many hate speech detection models is that mere mentions of identity terms result in false positive predictions \citep{dixon2018measuring}.

For a set of instances that contain mentions of identity terms we leverage HateCheck  \citep{rottger2021hatecheck}, which is a suite of functional tests targeted at diagnosing weaknesses of hate speech classifiers. Tests are constructed from hand-crafted templates, where the target is picked from a predefined set of protected groups. 


To evaluate our explanation method, we train six BERT classifiers on three different datasets \citep{founta2018large, vidgen2021introducing, davidson2017automated}, and obtain the necessity and sufficiency of the identity terms on explicitly hateful test cases in HateCheck that target \textit{women} and \textit{Muslims}. Details for the datasets and classifiers can be found in Appendix \ref{app:classifiers}.
We train classifiers on both `hate speech' labels and on more general `abusive' language labels in order to observe the effects of this distinction on the necessity and sufficiency values for identity terms. The difference between the two is that abusive language does not need to target a protected group identity. Accordingly, our first hypothesis is: 

\paragraph{\textit{Hypothesis 1:}}\textit{We will observe lower necessity for the identity terms for those models that are trained on the `abuse' labels compared to the ones that are trained on the `hate' labels.}  

\vspace{3pt}
\noindent We further hypothesize that different necessity and sufficiency scores for identity mentions in explicitly hateful test cases indicate different biases, and correlate with how well a model does in the HateCheck functionalities that aim to capture those biases (see Table \ref{tab:hatecheck-cases} for the test cases), leading to our second and third hypotheses:


\paragraph{\textit{Hypothesis 2:}} \textit{If a model has high sufficiency scores for identity terms in explicit expressions of hate (functionalities \textbf{F1}, \textbf{F2} and \textbf{F3}), this should indicate that the model is over-sensitive to identity terms. Therefore, we expect it have increased error rate in \textbf{F18} and \textbf{F19}, which consist of neutral or positive statements that mention identity terms.}

\paragraph{\textit{Hypothesis 3:}} \textit{If the necessity scores for identity terms are low in explicit expressions of hate, we can conclude under-reliance on the identity terms, and over-reliance on the context. Consequently, we expect that the model will perform poorly on \textbf{F22}, \textbf{F23} and \textbf{F24}, which capture abuse not targeted at protected identity groups. }

\begin{table*}[]
    \centering
    \small
    \begin{tabular}{r |c c | c c  }
    \toprule
    & \multicolumn{2}{c|}{\textbf{women}} & \multicolumn{2}{c}{\textbf{Muslims}} \\
    & \textit{necc} & \textit{suff}  & \textit{necc} & \textit{suff}  \\
    \midrule
        \textit{Founta2018-hate}  & 0.82 $\pm$0.18 & 0.29 $\pm$0.1 & 0.89 $\pm$0.16 & 0.81 $\pm$0.08  \\
         \textit{Founta2018-abuse} & 0.54 $\pm$0.17  & 0.34 $\pm$0.1 &  0.65 $\pm$0.21 & 0.82 $\pm$0.06 \\
         \textit{Davidson2017-hate} & 0.58 $\pm$0.09 & 0.21 $\pm$0.06  & 0.91 $\pm$0.12 & 0.74 $\pm$0.09 \\
         \textit{Davidson2017-abuse} & 0.82 $\pm$0.14 & 0.43 $\pm$0.13 & 0.83 $\pm$0.13 & 0.41 $\pm$0.14  \\
         \textit{Vidgen2021-hate} & 0.96 $\pm$0.02 & 0.71 $\pm$0.17  & 0.97 $\pm$0.03 & 0.88 $\pm$0.13  \\
         \textit{Vidgen2021-abuse} & 0.82 $\pm$0.14 & 0.64 $\pm$0.14 & 0.82 $\pm$0.15 & 0.88 $\pm$0.07 \\
    \bottomrule
    \end{tabular}
    \caption{The mean and standard deviation of necessity and sufficiency scores for target tokens in explicitly hateful cases of HateCheck (\textbf{F1}, \textbf{F2}, and \textbf{F3}) targeting \textbf{women} or \textbf{Muslims} for the the three classifiers trained on hate, and three classifiers trained on abuse labels. }
    \label{tab:hatecheck-necc-suff}
\end{table*}

\begin{table*}[]
    \centering
    \small
    \begin{tabular}{  r | >{\centering}p{2.5cm}  >{\centering}p{2cm}  | >{\centering}p{1.5cm}   >{\centering}p{1.5cm}   >{\centering\arraybackslash}p{1.5cm}  }
    \toprule
         & \multicolumn{2}{c |}{
         \begin{tabular}
         {@{}c@{}}
         \textit{Neutral/supportive group identity mention} \\ 
         \textbf{(F18, F19)}
         \end{tabular}} 
         & \multicolumn{3}{c }{
          \begin{tabular}
         {@{}c@{}}
         \textit{Abuse against non-protected targets} \\
         \textbf{(F22, F23, F24)}
         \end{tabular}
         } \\
         \cline{2-6} 
        \\[-1em]
         & \textbf{ women  } &  \textbf{Muslims} & 
         \textbf{group} & \textbf{individual} & \textbf{object} \\[2pt]

         \hline
         \textit{Founta2018-hate} & 0.02 & 0.78 & 0.19 & 0.15 & 0.05   \\
         \textit{Founta2018-abuse} & 0.02 & 0.78 & 0.45 & 0.72 & 0.37 \\
         \textit{Davidson2017-hate} & 0.02 & 0.78 & 0.37 & 0.18 & 0.02  \\
         \textit{Davidson2017-abuse} & 0.31 & 0.22 & 0.26 & 0.28 & 0.14  \\
         \textit{Vidgen2021-hate} & 0.36 & 0.82 & 0.02 & 0.00 & 0.00  \\
         \textit{Vidgen2021-abuse} & 0.42 & 0.96 & 0.40 & 0.61 & 0.00   \\
    \bottomrule
    \end{tabular}
    \caption{
    Proportions of test cases classified as hateful/abusive for different non-hateful HateCheck functionalities and targets. 
    }
    \label{tab:hatecheck-results}
\end{table*}

\subsection{Implementation} 

We obtain the average necessity and sufficiency values for explicitly hateful test cases targeting \textit{women} and \textit{Muslims} for each of the classifiers. 
We calculate necessity and sufficiency by masking a subset of the tokens and using our fine-tuned language model to generate infillings.
If multiple consecutive tokens are chosen, we aggregate them to a single mask instance to be infilled. 
We choose the number of perturbations for each example so that the expected number of perturbations for each token is 100. 
The necessity and sufficiency scores are only calculated for test cases that a classifier returns a correct prediction, since we only aim to explain positive predictions. The results can be found in Table~\ref{tab:hatecheck-necc-suff}. 
Table~\ref{tab:hatecheck-results} presents the proportions of test cases classified as hateful/abusive by each of the six classifiers on the non-hateful statements that mention identity terms (\textbf{F18} and \textbf{F19}) and abusive utterances not targeting protected identity groups (\textbf{F22}, \textbf{F23}, and \textbf{F24}). We report the results where necessity and sufficiency are calculated with masking rather than perturbing the chosen tokens in Appendix \ref{app:masking}.

As baselines, we calculate the average importance of the tokens corresponding to target groups with SHAP\footnote{\url{https://github.com/slundberg/shap}} and LIME\footnote{\url{https://github.com/marcotcr/lime}}. For both of these methods, we use the default parameters for textual data. As with the calculation of necessity and sufficiency, we only include the attribution scores for test cases on which the classifier correctly predicts the positive class. These results can be found in Table \ref{tab:hatecheck-lime-shap}. 

\section{Results and Discussion}

An example necessity and sufficiency attribution is given in Figure \ref{fig:example-necc-suff}. It shows that for this input, the token `Muslims' is more sufficient compared to `women', and the token `disgust' is more necessary in the context of `women' than that of `Muslims'.

According to our first hypothesis, we expect the models that were trained on the \textit{abuse} versions of each dataset to have lower necessity for identity terms compared to those that have been trained on \textit{hate} labels. Indeed, in Table \ref{tab:hatecheck-necc-suff} we observe this pattern for all models and targets except \textit{Davidson2017} for the target \textit{women}. This correctly suggests that identity terms are necessary for a comment to be hate speech, but not for it to be abusive. 

The results also clearly support our second hypothesis that if an identity mention has high sufficiency on explicit examples for a given model, then this model is over-sensitive to the identity term. Comparing the sufficiency of \textit{women} and \textit{Muslims} in Table~\ref{tab:hatecheck-necc-suff} illustrates this difference: for all models except \textit{Davidson2017-abuse} sufficiency is high for \textit{Muslims} and significantly lower for \textit{women}. Accordingly, all models except \textit{Davidson2017-abuse} display a large difference between their 
error rates on neutral or positive mentions for \textit{women} and \textit{Muslims} in Table~\ref{tab:hatecheck-results} (\textbf{F18}, \textbf{F19}). That is, the mere occurrence of the word ``Muslims'' is sufficient for the classifiers to classify a text as hate speech, even if the text is neutral. Furthermore within each group, higher sufficiency values correspond to higher error rates in functionalities \textbf{F18, F19}. \textit{Vidgen2021-hate} and \textit{Vidgen2021-abuse} display the highest sufficiency for \textit{women}, and correspondingly have the highest error rates on these test cases 
for \textit{women}. \textit{Davidson2017-abuse} has the lowest sufficiency for \textit{Muslims}, and the lowest error rate  
for this target.

Our third hypothesis is that low necessity for identity terms will be correlated with positive predictions for abusive instances that do not target a protected identity. In Table~\ref{tab:hatecheck-necc-suff}, the lowest necessity for both target groups are observed with \textit{Founta2018-abuse}. Indeed, this model has the highest 
rate of positive (abuse) predictions on all functionalities that test for abuse against non-protected targets in Table~\ref{tab:hatecheck-results}. The false positives in the test cases that target \textit{objects} is much higher than the corresponding errors for the other models, indicating that \textit{Founta2018-abuse} is indeed over-sensitive to abusive contexts, 
and does not consider the target of the abuse to be a necessary feature for the classification. 
On the other hand, the classifier trained on \textit{Vidgen2021-hate} shows the highest necessity values for both targets, and the lowest error rates on \textbf{F22}, \textbf{F23}, \textbf{F24}.

\begin{table*}[]
    \centering
    \small
    \begin{tabular}{r |c c | c c }
    \toprule
    & \multicolumn{2}{c|}{\textbf{women}} & \multicolumn{2}{c}{\textbf{Muslims}} \\
     & \textit{SHAP} & \textit{LIME} & \textit{SHAP} & \textit{LIME} \\
    \midrule
        \textit{Founta2018-hate} & 0.36 $\pm$0.24 & 0.018 $\pm$0.045 & 0.84 $\pm$0.2 & 0.013 $\pm$0.036 \\
         \textit{Founta2018-abuse} & -0.07 $\pm$0.21 & 0.011 $\pm$0.02 & 0.39 $\pm$0.25 & 0.042 $\pm$0.037 \\
         \textit{Davidson2017-hate} & 0.01 $\pm$0.02 & -0.021 $\pm$0.018 & 0.89 $\pm$0.15 & -0.002 $\pm$0.057 \\
         \textit{Davidson2017-abuse} & 0.42 $\pm$0.19 & 0.001 $\pm$0.086 & 0.37 $\pm$0.2 & -0.032 $\pm$0.092 \\
         \textit{Vidgen2021-hate} & 0.89 $\pm$0.17 & 0.012 $\pm$0.077 & 0.95 $\pm$0.11 & 0.042 $\pm$0.049 \\
         \textit{Vidgen2021-abuse}  & 0.66 $\pm$0.23 & 0.045 $\pm$0.088 & 0.75 $\pm$0.22 & 0.087 $\pm$0.062 \\
    \bottomrule
    \end{tabular}
    \caption{The mean and standard deviation of SHAP and LIME scores for target tokens in explicitly hateful cases of HateCheck (\textbf{F1}, \textbf{F2}, and \textbf{F3}) targeting \textbf{women} or \textbf{Muslims} for the the three classifiers trained on hate, and three classifiers trained on abuse labels. }
    \label{tab:hatecheck-lime-shap}
\end{table*}

\subsection{Comparison of Average SHAP and LIME Values with Necessity and Sufficiency} The average SHAP and LIME values for the two targets are presented in Table \ref{tab:hatecheck-lime-shap}. While \textit{Founta-abuse} and \textit{Davidson-abuse} get very similar SHAP scores for the target \textit{Muslims}, \textit{Founta2018-abuse} has high sufficiency for this token while \textit{Davidson2017-abuse} has high necessity. These two classifiers have very different false-positive rates for test instances that are non-abusive mentions of this target as reported in Table \ref{tab:hatecheck-results}, and hence can be observed to be biased against this group to a different extent. This distinction is clearly captured with the necessity and sufficiency scores, but not with SHAP. 

LIME scores seem even less consistent with the false-positive rates in Table \ref{tab:hatecheck-results} than SHAP. For example,  \textit{Davidson2017-hate} has a very high false-positive rate for neutral/supportive mentions of the target \textit{Muslims}, however the average LIME score for this model and target group is in low negatives. This means that LIME was unable to capture the biases of the model against this target group. 

For the target \textit{women}, LIME gives very similar average importance to the target tokens for \textit{Founta2018-abuse} and \textit{Vidgen2021-hate}, however all of the other metrics we calculate highlight significant differences. Indeed, we can observe in Table~\ref{tab:hatecheck-results} that \textit{Founta2018-abuse} has high false-positive rates for abuse against non-protected targets, but low false-positive rates for neutral/supportive mentions of the target \textit{women}, whereas \textit{Vidgen2021-hate} shows the exact opposite pattern. LIME scores seem unable to distinguish this important difference, unlike our necessity and sufficiency scores. 

\subsection{Qualitative analysis} 

Examining the generated perturbations reveals that the majority of the time, the generator successfully replaces the original snippet with non-offensive text. There are minimal repetitions of the same infill, indicating that the perturbations are sufficiently diverse. The quality of the perturbations is mixed, and tends to decrease with the number of infills. 
Our method can be used with any generator that can model the data distribution conditioned on the label. For our current 
model we prioritize diversity over fluency, and leave improvements on this aspect for future work. 

Inspecting the generated perturbations reveal that there are often instances that combine a few HateCheck functionalities. For example, as a perturbation for ``I hate {[}IDENTITY{]}''  model generates ``This is a subreddit I hate.'', combining shuffling the word order with changing the target. For the template ``{[}IDENTITY{]} are the dumbest people.'', one of the generated counterfactuals is ``I don't believe that American people are the dumbest people.'', 
combining negation and phrasing.

Interestingly, many infills for the identity terms do not clearly correspond to either object, individuals or non-protected groups, which are the three cases that HateCheck checks for. Instead, our generative model infills it with clauses such as ``I hate that I feel like I have to do these things'', actions such as ``I hate crying,'' or types of events like ``I hate surprises.'' This illustrates that our relatively non-constrained generation of counterfactuals provides better coverage of potential replacements, and provides a good complement to manual checks.

\section{Conclusion}
This work is a step towards more informative and transparent feature attribution metrics for explaining text classifiers. We argue that standard token importance metrics can be ambiguous in terms of what `importance' means. 
Instead, we adapt the theoretically-grounded concepts of \textit{necessity} and \textit{sufficiency} to explain text classifiers. 
Besides being more informative, the process of generating these two metrics is intuitive and can be explained to lay people in terms of ``how much the perturbations in input change the output of the classifier''. Moreover, the input perturbations can be presented to the users, leading to a transparent and understandable explainability framework. 

Considering the complexities of perturbing textual features, we introduced a practical implementation to compute the necessity and sufficiency of the input tokens. Taking hate speech detection as an example application, we showed that sufficiency and necessity can be used to explain the expected differences between a classifier that is intended to detect identity-based hate speech and those trained for detecting general abuse. We also leveraged these metrics to explain the observed over-sensitivity and under-sensitivity to mentions of target groups, issues that are tightly related to fairness in hate speech detection. 
While the current work focused on binary hate speech detection for English-language social media posts, 
in future work, we will explore the effectiveness of these metrics in generating explanations for other applications and languages. We will also explore how the new metrics can improve the debugging of the models or communicating the model’s decision-making process to the end-users.



\section{Ethical Considerations}

The proposed method has benefits and risks that should be considered from an ethics perspective.

One principle of ethical AI is \textit{transparency}, and we have developed this method with the goal of improving transparency for system developers, end users, and other stakeholders to better understand the inner workings of complex NLP systems. In the application domain of hate speech detection, we demonstrated how necessity and sufficiency scores might be used to diagnose 
possible classification biases against identity groups, who are frequently subjects of online abuse. This 
can help in addressing the known issue of over-sensitivity to identity terms, ensuring that benign conversations around issues concerning marginalized groups are not mis-classified as hate speech. 

However, there are also potential risks. We make use of existing datasets and thus our analysis is limited by those data: they were collected from public, online platforms without user's explicit consent, and may not accurately represent speakers from all demographic groups, they are only in English, and they may be biased towards or against certain topics of conversation. The data and analysis are also limited to the English language. Training language models on user data also has privacy implications, as the language model may then re-generate user text when deployed. 

While transparency and explainability are seen as desirable properties, they can also expose AI systems to malicious attacks. In the context of hate speech, our explainability metrics could potentially be used to identify and then exploit system vulnerabilities. 

Finally, our approach requires the use of large language models, which are computationally expensive to train and can reflect the biases of their training data. Our method of generating multiple counterfactual examples per word, rather than simply removing or masking that word, also increases the computational resources required.


    

\bibliography{anthology,custom}

\begin{thebibliography}{48}
\expandafter\ifx\csname natexlab\endcsname\relax\def\natexlab#1{#1}\fi

\bibitem[{Adadi and Berrada(2018)}]{adadi2018peeking}
Amina Adadi and Mohammed Berrada. 2018.
\newblock Peeking inside the black-box: {A} survey on explainable artificial
  intelligence {(XAI)}.
\newblock \emph{IEEE Access}, 6:52138--52160.

\bibitem[{Bahdanau et~al.(2015)Bahdanau, Cho, and Bengio}]{bahdanau2015neural}
Dzmitry Bahdanau, Kyung~Hyun Cho, and Yoshua Bengio. 2015.
\newblock Neural machine translation by jointly learning to align and
  translate.
\newblock In \emph{Proceedings of the 3rd International Conference on Learning
  Representations}.

\bibitem[{Borkan et~al.(2019)Borkan, Dixon, Sorensen, Thain, and
  Vasserman}]{borkan2019nuanced}
Daniel Borkan, Lucas Dixon, Jeffrey Sorensen, Nithum Thain, and Lucy Vasserman.
  2019.
\newblock Nuanced metrics for measuring unintended bias with real data for text
  classification.
\newblock In \emph{Companion Proceedings of the 2019 World Wide Web
  Conference}, pages 491--500.

\bibitem[{Chang et~al.(2018)Chang, Creager, Goldenberg, and
  Duvenaud}]{chang2018explaining}
Chun-Hao Chang, Elliot Creager, Anna Goldenberg, and David Duvenaud. 2018.
\newblock Explaining image classifiers by counterfactual generation.
\newblock In \emph{Proceedings of the International Conference on Learning
  Representations}.

\bibitem[{Danilevsky et~al.(2020)Danilevsky, Qian, Aharonov, Katsis, Kawas, and
  Sen}]{danilevsky2020survey}
Marina Danilevsky, Kun Qian, Ranit Aharonov, Yannis Katsis, Ban Kawas, and
  Prithviraj Sen. 2020.
\newblock \href {https://aclanthology.org/2020.aacl-main.46} {A survey of the
  state of explainable {AI} for natural language processing}.
\newblock In \emph{Proceedings of the 1st Conference of the Asia-Pacific
  Chapter of the Association for Computational Linguistics and the 10th
  International Joint Conference on Natural Language Processing}, pages
  447--459, Suzhou, China. Association for Computational Linguistics.

\bibitem[{Davidson et~al.(2017)Davidson, Warmsley, Macy, and
  Weber}]{davidson2017automated}
Thomas Davidson, Dana Warmsley, Michael Macy, and Ingmar Weber. 2017.
\newblock Automated hate speech detection and the problem of offensive
  language.
\newblock In \emph{Proceedings of the International AAAI Conference on Web and
  Social Media}, volume~11.

\bibitem[{Devlin et~al.(2019)Devlin, Chang, Lee, and
  Toutanova}]{devlin2019bert}
Jacob Devlin, Ming-Wei Chang, Kenton Lee, and Kristina Toutanova. 2019.
\newblock Bert: Pre-training of deep bidirectional transformers for language
  understanding.
\newblock In \emph{Proceedings of the Annual Conference of the North American
  Chapter of the Association for Computational Linguistics (NAACL-HLT)}.

\bibitem[{DeYoung et~al.(2020)DeYoung, Jain, Rajani, Lehman, Xiong, Socher, and
  Wallace}]{deyoung2020eraser}
Jay DeYoung, Sarthak Jain, Nazneen~Fatema Rajani, Eric Lehman, Caiming Xiong,
  Richard Socher, and Byron~C Wallace. 2020.
\newblock Eraser: A benchmark to evaluate rationalized {NLP} models.
\newblock In \emph{Proceedings of the 58th Annual Meeting of the Association
  for Computational Linguistics}, pages 4443--4458.

\bibitem[{Dixon et~al.(2018)Dixon, Li, Sorensen, Thain, and
  Vasserman}]{dixon2018measuring}
Lucas Dixon, John Li, Jeffrey Sorensen, Nithum Thain, and Lucy Vasserman. 2018.
\newblock Measuring and mitigating unintended bias in text classification.
\newblock In \emph{Proceedings of the 2018 AAAI/ACM Conference on AI, Ethics,
  and Society}, pages 67--73.

\bibitem[{Donahue et~al.(2020)Donahue, Lee, and Liang}]{donahue2020enabling}
Chris Donahue, Mina Lee, and Percy Liang. 2020.
\newblock Enabling language models to fill in the blanks.
\newblock In \emph{Proceedings of the 58th Annual Meeting of the Association
  for Computational Linguistics}, pages 2492--2501.

\bibitem[{Fong and Vedaldi(2017)}]{fong2017interpretable}
Ruth~C Fong and Andrea Vedaldi. 2017.
\newblock Interpretable explanations of black boxes by meaningful perturbation.
\newblock In \emph{Proceedings of the IEEE international conference on computer
  vision}, pages 3429--3437.

\bibitem[{Fortuna and Nunes(2018)}]{fortuna2018survey}
Paula Fortuna and S{\'e}rgio Nunes. 2018.
\newblock A survey on automatic detection of hate speech in text.
\newblock \emph{ACM Computing Surveys (CSUR)}, 51(4):1--30.

\bibitem[{Founta et~al.(2018)Founta, Djouvas, Chatzakou, Leontiadis, Blackburn,
  Stringhini, Vakali, Sirivianos, and Kourtellis}]{founta2018large}
Antigoni~Maria Founta, Constantinos Djouvas, Despoina Chatzakou, Ilias
  Leontiadis, Jeremy Blackburn, Gianluca Stringhini, Athena Vakali, Michael
  Sirivianos, and Nicolas Kourtellis. 2018.
\newblock Large scale crowdsourcing and characterization of {T}witter abusive
  behavior.
\newblock In \emph{Proceedings of the Twelfth International AAAI Conference on
  Web and Social Media}.

\bibitem[{Galhotra et~al.(2021)Galhotra, Pradhan, and
  Salimi}]{galhotra2021explaining}
Sainyam Galhotra, Romila Pradhan, and Babak Salimi. 2021.
\newblock Explaining black-box algorithms using probabilistic contrastive
  counterfactuals.
\newblock In \emph{Proceedings of the 2021 International Conference on
  Management of Data}, pages 577--590.

\bibitem[{Garg et~al.(2019)Garg, Perot, Limtiaco, Taly, Chi, and
  Beutel}]{garg2019counterfactual}
Sahaj Garg, Vincent Perot, Nicole Limtiaco, Ankur Taly, Ed~H Chi, and Alex
  Beutel. 2019.
\newblock Counterfactual fairness in text classification through robustness.
\newblock In \emph{Proceedings of the 2019 AAAI/ACM Conference on AI, Ethics,
  and Society}, pages 219--226.

\bibitem[{Guidotti et~al.(2018)Guidotti, Monreale, Ruggieri, Turini, Giannotti,
  and Pedreschi}]{Guidotti2018}
Riccardo Guidotti, Anna Monreale, Salvatore Ruggieri, Franco Turini, Fosca
  Giannotti, and Dino Pedreschi. 2018.
\newblock A survey of methods for explaining black box models.
\newblock \emph{ACM computing surveys (CSUR)}, 51(5):1--42.

\bibitem[{Halpern(2016)}]{halpern2016actual}
Joseph~Y. Halpern. 2016.
\newblock \emph{Actual Causality}.
\newblock MIT Press.

\bibitem[{Harbecke and Alt(2020)}]{harbecke2020considering}
David Harbecke and Christoph Alt. 2020.
\newblock Considering likelihood in {NLP} classification explanations with
  occlusion and language modeling.
\newblock In \emph{Proceedings of the 58th Annual Meeting of the Association
  for Computational Linguistics: Student Research Workshop}, pages 111--117.

\bibitem[{Hase et~al.(2021)Hase, Xie, and Bansal}]{hase2021the}
Peter Hase, Harry Xie, and Mohit Bansal. 2021.
\newblock The out-of-distribution problem in explainability and search methods
  for feature importance explanations.
\newblock \emph{Advances in Neural Information Processing Systems}, 34.

\bibitem[{Hooker et~al.(2019)Hooker, Erhan, Kindermans, and Kim}]{hooker2019a}
Sara Hooker, Dumitru Erhan, Pieter-Jan Kindermans, and Been Kim. 2019.
\newblock A benchmark for interpretability methods in deep neural networks.
\newblock \emph{Advances in Neural Information Processing Systems},
  32:9737--9748.

\bibitem[{Jacovi and Goldberg(2021)}]{jacovi2021aligning}
Alon Jacovi and Yoav Goldberg. 2021.
\newblock Aligning faithful interpretations with their social attribution.
\newblock \emph{Transactions of the Association for Computational Linguistics},
  9:294--310.

\bibitem[{Janzing et~al.(2020)Janzing, Minorics, and
  Bl{\"o}baum}]{janzing2020feature}
Dominik Janzing, Lenon Minorics, and Patrick Bl{\"o}baum. 2020.
\newblock Feature relevance quantification in explainable {AI}: A causal
  problem.
\newblock In \emph{Proceedings of the International Conference on Artificial
  Intelligence and Statistics}, pages 2907--2916. PMLR.

\bibitem[{Kaushik et~al.(2021)Kaushik, Setlur, Hovy, and
  Lipton}]{kaushik2021explaining}
Divyansh Kaushik, Amrith Setlur, Eduard~H Hovy, and Zachary~Chase Lipton. 2021.
\newblock Explaining the efficacy of counterfactually augmented data.
\newblock In \emph{Proceedings of the International Conference on Learning
  Representations}.

\bibitem[{Kim et~al.(2020)Kim, Yi, Kim, and Yoon}]{kim2020interpretation}
Siwon Kim, Jihun Yi, Eunji Kim, and Sungroh Yoon. 2020.
\newblock \href {https://doi.org/10.18653/v1/2020.emnlp-main.255}
  {Interpretation of {NLP} models through input marginalization}.
\newblock In \emph{Proceedings of the 2020 Conference on Empirical Methods in
  Natural Language Processing (EMNLP)}, pages 3154--3167, Online. Association
  for Computational Linguistics.

\bibitem[{Kiritchenko et~al.(2021)Kiritchenko, Nejadgholi, and
  Fraser}]{kiritchenko2021confronting}
Svetlana Kiritchenko, Isar Nejadgholi, and Kathleen~C Fraser. 2021.
\newblock Confronting abusive language online: A survey from the ethical and
  human rights perspective.
\newblock \emph{Journal of Artificial Intelligence Research}, 71:431--478.

\bibitem[{Lundberg and Lee(2017)}]{lundberg2017unified}
Scott~M Lundberg and Su-In Lee. 2017.
\newblock \href
  {http://papers.nips.cc/paper/7062-a-unified-approach-to-interpreting-model-predictions.pdf}
  {A unified approach to interpreting model predictions}.
\newblock In I.~Guyon, U.~V. Luxburg, S.~Bengio, H.~Wallach, R.~Fergus,
  S.~Vishwanathan, and R.~Garnett, editors, \emph{Advances in Neural
  Information Processing Systems 30}, pages 4765--4774. Curran Associates, Inc.

\bibitem[{Luo et~al.(2021)Luo, Ivison, Han, and Poon}]{luo2021local}
Siwen Luo, Hamish Ivison, Caren Han, and Josiah Poon. 2021.
\newblock Local interpretations for explainable natural language processing: A
  survey.
\newblock \emph{arXiv preprint arXiv:2103.11072}.

\bibitem[{Mathew et~al.(2021)Mathew, Saha, Yimam, Biemann, Goyal, and
  Mukherjee}]{mathew2021hatexplain}
Binny Mathew, Punyajoy Saha, Seid~Muhie Yimam, Chris Biemann, Pawan Goyal, and
  Animesh Mukherjee. 2021.
\newblock Hatexplain: A benchmark dataset for explainable hate speech
  detection.
\newblock In \emph{Proceedings of the AAAI Conference on Artificial
  Intelligence}, volume~35, pages 14867--14875.

\bibitem[{Miller(2019)}]{miller2019explanation}
Tim Miller. 2019.
\newblock Explanation in artificial intelligence: {I}nsights from the social
  sciences.
\newblock \emph{Artificial intelligence}, 267:1--38.

\bibitem[{Mothilal et~al.(2021)Mothilal, Mahajan, Tan, and
  Sharma}]{mothilal2021towards}
Ramaravind Mothilal, Divyat Mahajan, Chenhao Tan, and Amit Sharma. 2021.
\newblock Towards unifying feature attribution and counterfactual explanations:
  {D}ifferent means to the same end.
\newblock In \emph{Proceedings of the 2021 AAAI/ACM Conference on AI, Ethics,
  and Society}, pages 652--663.

\bibitem[{Nejadgholi and Kiritchenko(2020)}]{nejadgholi2020cross}
Isar Nejadgholi and Svetlana Kiritchenko. 2020.
\newblock On cross-dataset generalization in automatic detection of online
  abuse.
\newblock In \emph{Proceedings of the Fourth Workshop on Online Abuse and
  Harms}, pages 173--183.

\bibitem[{Park et~al.(2018)Park, Shin, and Fung}]{park2018reducing}
Ji~Ho Park, Jamin Shin, and Pascale Fung. 2018.
\newblock Reducing gender bias in abusive language detection.
\newblock In \emph{Proceedings of the Conference on Empirical Methods in
  Natural Language Processing}, pages 2799--2804, Brussels, Belgium.

\bibitem[{Pearl(2009)}]{pearl2009causality}
Judea Pearl. 2009.
\newblock \emph{Causality}.
\newblock Cambridge University Press.

\bibitem[{Pham et~al.(2021)Pham, Bui, Mai, and Nguyen}]{pham2021double}
Thang~M Pham, Trung Bui, Long Mai, and Anh Nguyen. 2021.
\newblock Double trouble: How to not explain a text classifier's decisions
  using counterfactuals synthesized by masked language models?
\newblock \emph{arXiv preprint arXiv:2110.11929}.

\bibitem[{Radford et~al.(2019)Radford, Wu, Child, Luan, Amodei, Sutskever
  et~al.}]{radford2019language}
Alec Radford, Jeffrey Wu, Rewon Child, David Luan, Dario Amodei, Ilya
  Sutskever, et~al. 2019.
\newblock Language models are unsupervised multitask learners.
\newblock \emph{OpenAI blog}, 1(8):9.

\bibitem[{Ribeiro et~al.(2016)Ribeiro, Singh, and Guestrin}]{ribeiro2016should}
Marco~Tulio Ribeiro, Sameer Singh, and Carlos Guestrin. 2016.
\newblock ``{W}hy should i trust you?'' {E}xplaining the predictions of any
  classifier.
\newblock In \emph{Proceedings of the 22nd ACM SIGKDD International Conference
  on Knowledge Discovery and Data Mining}, pages 1135--1144.

\bibitem[{Ribeiro et~al.(2020)Ribeiro, Wu, Guestrin, and
  Singh}]{ribeiro2020beyond}
Marco~Tulio Ribeiro, Tongshuang Wu, Carlos Guestrin, and Sameer Singh. 2020.
\newblock \href {https://doi.org/10.18653/v1/2020.acl-main.442} {Beyond
  accuracy: Behavioral testing of {NLP} models with {C}heck{L}ist}.
\newblock In \emph{Proceedings of the 58th Annual Meeting of the Association
  for Computational Linguistics}, pages 4902--4912, Online. Association for
  Computational Linguistics.

\bibitem[{Ross et~al.(2021)Ross, Marasovi{\'c}, and
  Peters}]{ross2021explaining}
Alexis Ross, Ana Marasovi{\'c}, and Matthew Peters. 2021.
\newblock \href {https://doi.org/10.18653/v1/2021.findings-acl.336} {Explaining
  {NLP} models via minimal contrastive editing ({M}i{CE})}.
\newblock In \emph{Findings of the Association for Computational Linguistics:
  ACL-IJCNLP 2021}, pages 3840--3852, Online. Association for Computational
  Linguistics.

\bibitem[{Ross et~al.(2020)Ross, Marasovi{\'c}, and
  Peters}]{ross2020explaining}
Alexis Ross, Ana Marasovi{\'c}, and Matthew~E Peters. 2020.
\newblock Explaining {NLP} models via minimal contrastive editing {(MiCE)}.
\newblock \emph{arXiv preprint arXiv:2012.13985}.

\bibitem[{R{\"o}ttger et~al.(2021)R{\"o}ttger, Vidgen, Nguyen, Waseem,
  Margetts, and Pierrehumbert}]{rottger2021hatecheck}
Paul R{\"o}ttger, Bertie Vidgen, Dong Nguyen, Zeerak Waseem, Helen Margetts,
  and Janet Pierrehumbert. 2021.
\newblock \href {https://doi.org/10.18653/v1/2021.acl-long.4} {{H}ate{C}heck:
  Functional tests for hate speech detection models}.
\newblock In \emph{Proceedings of the 59th Annual Meeting of the Association
  for Computational Linguistics and the 11th International Joint Conference on
  Natural Language Processing (Volume 1: Long Papers)}, pages 41--58, Online.
  Association for Computational Linguistics.

\bibitem[{Sundararajan et~al.(2017)Sundararajan, Taly, and
  Yan}]{sundararajan2017axiomatic}
Mukund Sundararajan, Ankur Taly, and Qiqi Yan. 2017.
\newblock Axiomatic attribution for deep networks.
\newblock In \emph{Proceedings of the International Conference on Machine
  Learning}, pages 3319--3328. PMLR.

\bibitem[{Vaughan and Wallach(2021)}]{vaughan2021human}
Jennifer~Wortman Vaughan and Hanna Wallach. 2021.
\newblock A human-centered agenda for intelligible machine learning.
\newblock In Marcello Pelillo and Teresa Scantamburlo, editors, \emph{Machines
  We Trust: Perspectives on Dependable AI}. The MIT Press.

\bibitem[{Vidgen et~al.(2021)Vidgen, Nguyen, Margetts, Rossini, and
  Tromble}]{vidgen2021introducing}
Bertie Vidgen, Dong Nguyen, Helen Margetts, Patricia Rossini, and Rebekah
  Tromble. 2021.
\newblock Introducing {CAD}: the contextual abuse dataset.
\newblock In \emph{Proceedings of the 2021 Conference of the North American
  Chapter of the Association for Computational Linguistics: Human Language
  Technologies}, pages 2289--2303.

\bibitem[{Watson et~al.(2021)Watson, Gultchin, Taly, and
  Floridi}]{watson2021local}
David Watson, Limor Gultchin, Ankur Taly, and Luciano Floridi. 2021.
\newblock Local explanations via necessity and sufficiency: {U}nifying theory
  and practice.
\newblock In \emph{Proceedings of Machine Learning Research}, volume 161, page
  1382–1392.

\bibitem[{Wu et~al.(2021)Wu, Ribeiro, Heer, and Weld}]{wu2021polyjuice}
Tongshuang Wu, Marco~Tulio Ribeiro, Jeffrey Heer, and Daniel Weld. 2021.
\newblock \href {https://doi.org/10.18653/v1/2021.acl-long.523} {Polyjuice:
  Generating counterfactuals for explaining, evaluating, and improving models}.
\newblock In \emph{Proceedings of the 59th Annual Meeting of the Association
  for Computational Linguistics and the 11th International Joint Conference on
  Natural Language Processing (Volume 1: Long Papers)}, pages 6707--6723,
  Online. Association for Computational Linguistics.

\bibitem[{Wulczyn et~al.(2017)Wulczyn, Thain, and Dixon}]{wulczyn2017ex}
Ellery Wulczyn, Nithum Thain, and Lucas Dixon. 2017.
\newblock Ex machina: Personal attacks seen at scale.
\newblock In \emph{Proceedings of the 26th International Conference on World
  Wide Web}, pages 1391--1399.

\bibitem[{Zaidan et~al.(2007)Zaidan, Eisner, and Piatko}]{zaidan2007using}
Omar Zaidan, Jason Eisner, and Christine Piatko. 2007.
\newblock Using “annotator rationales” to improve machine learning for text
  categorization.
\newblock In \emph{Proceedings of the Conference of the North American Chapter
  of the Association for Computational Linguistics}, pages 260--267.

\bibitem[{Zhang et~al.(2019)Zhang, Baldridge, and He}]{zhang2019paws}
Yuan Zhang, Jason Baldridge, and Luheng He. 2019.
\newblock Paws: Paraphrase adversaries from word scrambling.
\newblock In \emph{Proceedings of the 2019 Conference of the North American
  Chapter of the Association for Computational Linguistics: Human Language
  Technologies, Volume 1 (Long and Short Papers)}, pages 1298--1308.

\end{thebibliography}
\bibliographystyle{acl_natbib}

\newpage
\appendix


\section{Data, Training and Generation Details for the Infilling Language Model}
\label{app:ILM}

To fine-tune the ILM model, we use the following four datasets: \textit{Wikipedia Toxicity}\footnote{\url{https://figshare.com/articles/dataset/Wikipedia_Talk_Labels_Toxicity/4563973}} \cite{wulczyn2017ex}, \textit{Founta2018}\footnote{\url{https://github.com/ENCASEH2020/hatespeech-twitter}} \cite{founta2018large}, \textit{Civil Comments}
\footnote{\url{https://bit.ly/3Kfaveb}}
\cite{borkan2019nuanced}, and \textit{Vidgen2021}\footnote{\url{https://zenodo.org/record/4881008\#.YeBBQ2jMKUk}} \cite{vidgen2021introducing}. 
The datasets contain English-language utterances, and cover different domains (Twitter post, Reddit posts, Wikipedia comments, and comments from news websites). 
The datasets have been created to study abusive language, and are commonly used to train and evaluate classification models that detect various sub-categories of online abuse, such as hate speech, toxicity, personal attacks, etc. 
All datasets except \textit{Founta2018} are in the public domain and licensed for research purposes. \textit{Founta2018} dataset is being used with the permission of the first author. 

The details on each dataset are provided in Table~\ref{tab:data-generative-model}. 
For the \textit{Wikipedia Toxicity} dataset, a large portion of the data is from conversations about Wikipedia-specific topics. To not skew our generation model, we filter these instances following the unsupervised method presented by \citet{nejadgholi2020cross}\footnote{\url{https://github.com/IsarNejad/cross_dataset_toxicity}}. Because the \textit{Civil Comments} dataset is significantly larger than the rest, we randomly sample 30K neutral instances and discard the rest. After filtering, the compound dataset of neutral instances consists of 130,430 instances in total. As preprocessing, we replace URLs, mentions and emojis with special tokens.

To train the ILM, we fine-tune GPT-2 (1.5B parameters) for 4 epochs with the default hyper-parameters provided by \citet{donahue2020enabling}. The training takes approximately 2.5 hours on a Tesla V100-SXM2 GPU. Although the original ILM is trained by infilling words, $n$-grams, sentences and paragraphs, we modify the objective to only infill words and $n$-grams. 

We generate perturbations once for the 120 HateCheck cases, and evaluate all models on the same set of perturbations. The number of perturbations are chosen so that to have approximately 100 perturbed instances for each token for the necessity calculation, and 100 instances for the sufficiency calculation. This results in a total of 66,120 perturbed instances, and takes approximately 6 hours to generate on a 2.3 GHz Quad-Core Intel Core i7 CPU.

\setcounter{table}{0}
\renewcommand\thetable{A.\arabic{table}}

\setlength{\tabcolsep}{4pt}
\begin{table}[t]
    \centering
    \small
    \begin{tabular}{ p{2.8cm} p{1.8cm} p{1cm} p{1cm}  }
    \hline
    \textbf{Dataset} & \textbf{Source} & \textbf{Class}&\textbf{Size } \\
    \hline
 \makecell[tl]{Wikipedia Toxicity  \\\cite{wulczyn2017ex}} & \makecell[tl]{Wikipedia\\comments} &Normal& 36,121 \\
 \hline
  \makecell[tl]{Founta2018\\ \cite{founta2018large} }& \makecell[tl]{Twitter\\posts} &Normal&  53,236 \\ [2pt]
 \hline 
  \makecell[tl]{Civil Comments \\\cite{borkan2019nuanced}}& \makecell[tl]{Comments on\\ news sites } & Normal&30,000\\ [2pt]
  
    \hline 
  \makecell[tl]{Vidgen2021 \\\cite{vidgen2021introducing}}& \makecell[tl]{Reddit\\ posts} & Non-Abusive& 11,073 \\ [5pt]
 \hline
 \textbf{Total} & & & \textbf{130,430} \\
\hline
    \end{tabular}
    \caption{Description of the training data used to fine-tune the ILM model.}
    \label{tab:data-generative-model}
\end{table}
\setlength{\tabcolsep}{6pt}

\setcounter{table}{0}
\renewcommand\thetable{B.\arabic{table}}

\setlength{\tabcolsep}{4pt}
\begin{table*}[tbh]
    \centering
    \small
    \begin{tabular}{ p{2.5cm} p{2.8cm} p{2cm} p{3.5cm} r r r }
    \hline
    \textbf{Classifier} & \textbf{Dataset} & \textbf{Positive Class }& \textbf{Negative Class} & \multicolumn{3}{c}{\textbf{Size }} \\
    & \textbf{} & & & \textit{Train} & \textit{Dev} & \textit{Test} \\
    \hline
   \makecell[tl]{\textit{Founta2018-hate}} & \makecell[tl]{Founta2018\\ \cite{founta2018large} }& Hateful &  \makecell[tl]{Normal\\Abusive} & \makecell{62,445} & \makecell{7,806} &\makecell{7,806}\\
 \hline
   \makecell[tl]{\textit{Founta2018-abuse}} & \makecell[tl]{Founta2018\\ \cite{founta2018large} }&  \makecell[tl]{Hateful\\Abusive} &  Normal& \makecell{62,445} & \makecell{7,806} &\makecell{7,806}\\
  \hline 
 \makecell[tl]{\textit{Davidson2017-hate}} & \makecell[tl]{Davidson2017 \\\cite{davidson2017automated}}&  Hate   &  \makecell[tl]{Neither\\Offensive} & \makecell{19,826} & \makecell{2,478} &\makecell{2,479}\\
  \hline 
 \makecell[tl]{\textit{Davidson2017-abuse}} & \makecell[tl]{Davidson2017 \\\cite{davidson2017automated}}&  \makecell[tl]{Hate\\Offensive}   &  \makecell[tl]{Neither} & \makecell{19,826} &\makecell{2,478} & \makecell{2,479}\\
  \hline 
 \makecell[tl]{\textit{Vidgen2021-hate}} & \makecell[tl]{Vidgen2021 \\\cite{vidgen2021introducing}}&   Identity-directed abuse & \makecell[tl]{Non-abusive\\
 Person-directed abuse\\
Affiliation-directed abuse} & \makecell{13,585}  & \makecell{4,527} &\makecell{5,308}\\
 \hline 
 \makecell[tl]{\textit{Vidgen2021-abuse}} & \makecell[tl]{Vidgen2021 \\\cite{vidgen2021introducing}}&   Abusive & Non-abusive  & \makecell{13,585} & \makecell{4,527} &\makecell{5,308}\\
 \hline
 
\hline
    \end{tabular}
    \caption{Description of the datasets used to fine-tune hate speech and abuse detection classifiers. }
    \label{tab:data-classifiers}
\end{table*}
\setlength{\tabcolsep}{6pt}

\begin{table}[tbh]
\small
    \centering
    \begin{tabular}{ l | c | c | c   }
    \hline
    & Micro & Macro & Training time \\
    & F1 & F1 &  (mins) \\
    \hline
         \textit{Founta2018-hate} & 0.94 & 0.67 &  28   \\
         \textit{Founta2018-abuse} & 0.94 & 0.93 &  28 \\
         \textit{Davidson2017-hate} & 0.94 & 0.70 &  7 \\
         \textit{Davidson2017-abuse} & 0.96 & 0.93 & 7  \\
         \textit{Vidgen2021-hate} & 0.91 & 0.71 & 22  \\
         \textit{Vidgen2021-abuse} & 0.85 & 0.72 &  22  \\
    \hline
    \end{tabular}
    \caption{Micro- and macro-averaged F1-scores and training times for each BERT model trained and evaluated on the given datasets. 
    }
    \label{tab:classifier-acc}
\end{table}

\section{Data and Training Details for Hate Speech Classifiers}
\label{app:classifiers}


We fine-tune six BERT \citep{devlin2019bert} classifiers on three different datasets and with two different labelling schemes (hate speech vs. abusive language) for each. 
The datasets include: \textit{Davidson2017}\footnote{\url{https://github.com/t-davidson/hate-speech-and-offensive-language/tree/master/data}}  \cite{davidson2017automated}, \textit{Founta2018} \cite{founta2018large}, and \textit{Vidgen2021} \cite{vidgen2021introducing}. 
The datasets contain English-language posts from two online platforms, Twitter and Reddit. 
The details on each dataset are provided in Table~\ref{tab:data-classifiers}. 

We train two models on the dataset of \citet{founta2018large}. For \textit{Founta2018-hate}, we binarize the labels to map hate annotations as positive, and the rest as the negative class. For \textit{Founta2018-abuse}, we label both hate and abuse annotations as positive, and the rest as negative. To illustrate that our method can provide explanations for models trained on data that is not explicitly modelled by our perturbation generator, we also train models on two versions of the dataset of \citet{davidson2017automated}: \textit{Davidson2017-abuse} and \textit{Davidson2017-hate}, which are binarized in the same manner.

The dataset of \citet{vidgen2021introducing} provides a hierarchical labelling scheme, the top distinction being \textit{abusive} vs. \textit{non-abusive}. We binarize \textit{Vidgen2021-abuse} based on these labels. For \textit{Vidgen2021-hate}, we take the positive class to be those instances that are labelled \textit{identity-directed abuse}, and label the rest as the negative class. 

We employ the same pre-processing steps as in the experiments by \citet{rottger2021hatecheck}, and replace URLs, mentions and emojis with special tokens. We fine-tune a BERT model from the Hugging Face library\footnote{\url{https://huggingface.co/bert-base-uncased}} on each of these datasets on a single Tesla V100-SXM2 GPU. Each model has 110M trainable parameters. We follow the implementation of \citet{rottger2021hatecheck} and use their hyper-parameters of 3 epochs, batch size of 16, learning rate of 5e-5 and weight decay of 0.01. We also employ weighted cross-entropy loss that corrects for the class imbalance in data. For the training/development/test splits, we use the standard split for \textit{Vidgen2021} provided by the creators of the dataset, and use a stratified 80/10/10 split for the other datasets, making sure that the splits are the same for the \textit{hate} and \textit{abuse} versions of each, and correspond to the training set for ILM when applicable. The classification performance of these models on the held-out test sets is shown in Table \ref{tab:classifier-acc}, together with the training times for each.
We can observe that the reported scores are within a few percentage points of the previously published results \citep{rottger2021hatecheck}. All reported results are from a single run.

\begin{table}[]
    \centering
    \small
    \begin{tabular}{r |c c | c c}
    \toprule
    & \multicolumn{2}{c|}{\textbf{women}} & \multicolumn{2}{c}{\textbf{Muslims}} \\
    & \textit{necc} & \textit{suff} & \textit{necc} & \textit{suff} \\
    \midrule
        \textit{Founta2018-hate}  & 0.53 & 0.30 & 0.72 & 0.81   \\
         \textit{Founta2018-abuse} & 0.19 & 0.34 & 0.36 & 0.82  \\
         \textit{Davidson2017-hate} & 0.44 & 0.21 & 0.88 & 0.74 \\
         \textit{Davidson2017-abuse} & 0.55 & 0.44 & 0.52 & 0.41  \\
         \textit{Vidgen2021-hate} & 0.87 & 0.71 & 0.93 & 0.88 \\
         \textit{Vidgen2021-abuse} & 0.62 & 0.64 & 0.64 & 0.88 \\
    \bottomrule
    \end{tabular}
    \caption{Average necessity and sufficiency scores calculated by masking rather than perturbing selected tokens, for the identity terms in explicitly hateful cases of HateCheck (\textbf{F1}, \textbf{F2}, and \textbf{F3}) targeting \textbf{women} or \textbf{Muslims} for the the three classifiers trained on hate, and three classifiers trained on abuse labels. }
    \label{tab:masked-necc-suff}
\end{table}

\section{Calculating Necessity and Sufficiency with Masking}\label{app:masking}

In Section \ref{sec:introduction} we have argued that using the \textit{mask} token from the pre-training objective in feature attribution methods has several drawbacks. Nevertheless, in Table \ref{tab:masked-necc-suff} we report the results of a modified version of our experiment presented in Section \ref{sec:experiments} where we keep the number and the location of the perturbations the same as the original experiments, but instead of perturbing the chosen tokens using an LM, we replace them with the \textit{mask} token. The results show that although the values are different than their counterparts in the main experiment, the overall trends remain the same, and support the hypotheses presented in Section \ref{sec:experiments}. Evaluating the classifier with the masked input is faster than explicitly generating perturbations, but the method ceases to be model agnostic and looses transparency. The results still suggest that evaluating necessity and sufficiency with masked rather than perturbed inputs might be preferable in contexts where latency is more important than transparency, or as a pre-processing step to choose which inputs and tokens to focus on for in-depth analysis with explicit perturbations. We leave further explorations of this avenue for future work.

\end{document}